\documentclass[conference]{IEEEtran}
\usepackage{amsmath,amssymb,booktabs,multirow,array,graphicx,xcolor,tabularx}
\usepackage{hyperref,url,enumitem}
\hypersetup{colorlinks=true,linkcolor=blue,citecolor=blue,urlcolor=blue}
\newcommand{\sas}{\textsc{SASGeo}}
\newcommand{\yes}{\(\checkmark\)}
\newcommand{\no}{--}
\title{SASGeo: Stability-Aware Semantic Map Localization for GNSS-Denied UAVs -- A Framework and Synthetic Proof of Concept}
\author{\IEEEauthorblockN{Natalia Trukhina and Vadim Vashkelis}\IEEEauthorblockA{Embedded Intelligence Lab (emilab.org)\\\{ntrukhina, vvashkelis\}@emilab.org}}
\begin{document}
\maketitle
\begin{abstract}
GNSS-denied unmanned aerial vehicles require occasional absolute position fixes to bound the drift of visual--inertial odometry. Cross-view image retrieval can provide such fixes, but raw appearance is sensitive to season, illumination, viewpoint, map age, and sensor modality. We propose \sas, a semantic map-localization framework that represents the environment through persistent structures such as roads, buildings, waterways, railways, intersections, and field boundaries. The method combines semantic raster alignment, relational graph evidence, feature stability and geographic distinctiveness, explicit positive/contradictory/unknown observations, and integrity-aware rejection of ambiguous fixes. Unlike a broad architecture-only proposal, this paper specifies concrete weighting and decision models and reports a reproducible synthetic proof of concept. In 220 randomized retrieval trials with rotation, scale changes, partial crops, occlusion, simulated map changes, and hard semantic decoys, a global semantic descriptor achieved 58.6\% Recall@1, while spatial semantic matching variants achieved 94.5--95.5\%. Wilson 95\% intervals separate the global descriptor from the spatial variants but overlap among the spatial variants, so the experiment supports semantic geometry rather than a definitive benefit from each proposed module. The preliminary experiment does not validate real-flight navigation; rather, it demonstrates that structured semantic geometry can discriminate locations under controlled cross-view perturbations and identifies the harder aliasing, map-aging, and rejection tests required next.
\end{abstract}

\section{Introduction}
GNSS is lightweight and globally referenced, but can become unreliable under multipath, obstruction, jamming, or spoofing. Visual--inertial odometry (VIO) supplies high-rate relative motion, yet its error grows without absolute observations \cite{qin2018vins}. A UAV can obtain global corrections by matching onboard imagery to georeferenced satellite or aerial imagery. Learned cross-view retrieval has progressed through datasets such as University-1652 and SUES-200 and practical UAV--satellite matching methods \cite{zheng2020university,zhu2023sues,ding2021practical}, but image appearance can differ substantially with altitude, season, illumination, viewpoint, sensor modality, and map age.

Many geographic structures are more persistent than their pixels. Road topology, rivers, railways, bridges, building footprints, and coastlines may remain identifiable even when color and texture change. Prior studies have consequently explored OpenStreetMap (OSM), vector maps, semantic embeddings, object graphs, road-geometry bird's-eye-view (BEV) calibration, and sequential filters \cite{vojir2020semantic,zhou2021osm,schmidt2025osmselflocalization,zilke2025semanticmapdata,ouyang2024vector,hu2024osm,wang2025vecmaplocnet,popov2026trafficcalibration,zhang2025hierarchical,li2025graph,chen2025swapf}. However, the literature generally covers subsets of five requirements that matter for safety-critical map fixes: dense semantic evidence, relational verification, temporal consistency, explicit persistence modeling, and a calibrated option to reject ambiguous observations.

\begin{figure*}[t]
\centering\includegraphics[width=\textwidth]{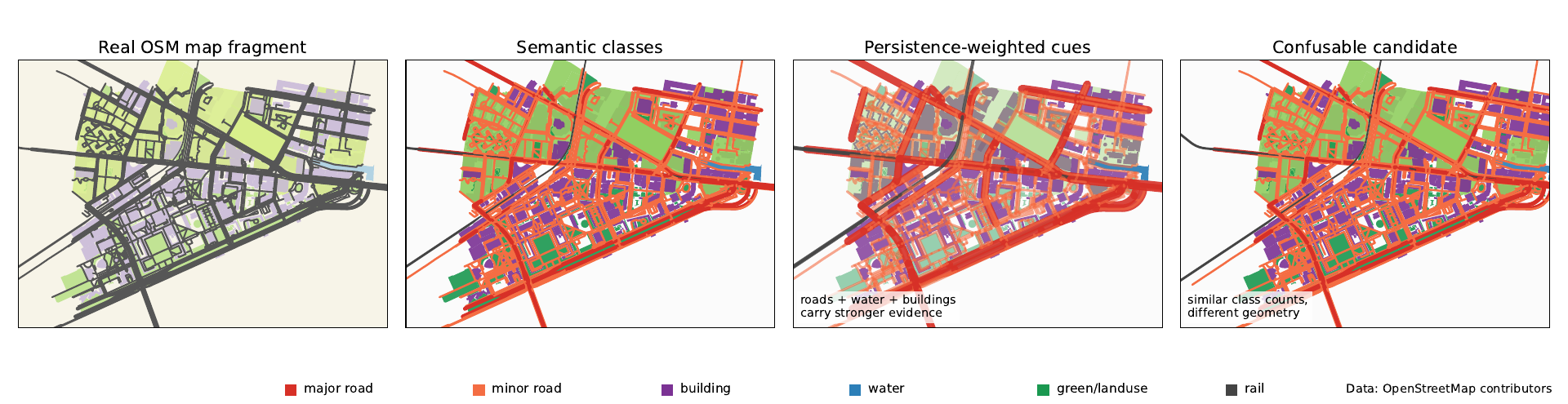}
\caption{Reader-level intuition for \sas{} using a real OpenStreetMap vector extract near Cambridge, Massachusetts. A map fragment is converted into semantic classes such as roads, buildings, water, green/land-use areas, and railways. Persistent and distinctive structures receive stronger evidence weights, while a confusable candidate can share similar class counts but differ in semantic geometry. Data: \copyright{} OpenStreetMap contributors, ODbL; rendered locally from vector features rather than from a map tile.}
\label{fig:intuition}
\end{figure*}

This paper proposes \sas, whose technical thesis is that an absolute map fix should be based on the joint evidence of (i) dense semantic alignment, (ii) semantic-object relationships, (iii) temporal and map-age persistence, and (iv) an integrity decision that can withhold the fix. The contributions are:
\begin{itemize}[leftmargin=*]
\item an operational stability-and-distinctiveness model rather than an unspecified semantic weight;
\item a joint raster--graph--temporal localization objective with separate positive, contradictory, and unknown evidence;
\item a decision-theoretic acceptance rule and risk--coverage evaluation for false absolute fixes;
\item a reproducible synthetic proof of concept with controlled perturbations, hard decoys, confidence intervals, and ablated risk--coverage curves;
\item an embedded implementation and real-world evaluation plan that clearly distinguishes design targets from measured results.
\end{itemize}
The experiment is intentionally modest. It tests the central semantic-discrimination mechanism, not end-to-end flight performance, learned perception, or calibrated field integrity.

\section{Related Work and Positioning}
\subsection{Cross-view and semantic map localization}
Cross-view methods commonly retrieve a satellite tile from a UAV image and optionally refine the pose. University-1652, SUES-200, and location-classification matching established strong benchmark-driven formulations for UAV--satellite retrieval \cite{zheng2020university,zhu2023sues,ding2021practical}. Hierarchical matching adds semantic or structural constraints before fine correspondences \cite{zhang2025hierarchical}. Semantic-map methods instead align camera observations with vector or rasterized geographic information: Vojir et al. used building instances \cite{vojir2020semantic}; Zhou et al. used image-to-OSM likelihoods with sequential Monte Carlo \cite{zhou2021osm}; Schmidt et al. and Zilke matched semantic UAV observations to OSM/map data \cite{schmidt2025osmselflocalization,zilke2025semanticmapdata}; Ouyang et al. combined vector-shape matching and particle filtering \cite{ouyang2024vector}; and VecMapLocNet learned UAV-to-vector-map features \cite{wang2025vecmaplocnet}. Object-graph matching and semantic-weighted particle filtering are recent complementary directions \cite{li2025graph,chen2025swapf}.

\begin{table*}[t]
\caption{Positioning relative to representative semantic, vector-map, and cross-view UAV localization work. ``Persist.'' denotes explicit persistence modeling under map age, season, visibility, or transient content; ``reject'' denotes an explicit decision to withhold an uncertain absolute fix. \sas{} entries are proposed capabilities, not real-flight validation claims.}
\label{tab:comparison}
\centering\footnotesize
\setlength{\tabcolsep}{3.5pt}
\begin{tabularx}{\textwidth}{p{2.05cm}Xcccccc}
\toprule
Method & Main reference cue & Dense & Graph & Temporal & Persist. & Reject & Unknown \\
\midrule
Vojir et al. \cite{vojir2020semantic} & OSM building instances & partial & partial & \no & \no & \no & \no \\
Zhou et al. \cite{zhou2021osm} & OSM semantic likelihood & \yes & \no & SMC & \no & partial & \no \\
Ouyang et al. \cite{ouyang2024vector} & Vector-map shapes & partial & partial & PF & \no & partial & \no \\
Hu et al. \cite{hu2024osm} & OSM with satellite imagery & \yes & \no & \no & \no & \no & \no \\
VecMapLoc. \cite{wang2025vecmaplocnet} & UAV-to-vector-map features & \yes & partial & \no & \no & \no & \no \\
Zhang et al. \cite{zhang2025hierarchical} & Satellite coarse-to-fine matching & \yes & partial & \no & \no & margin & \no \\
Liu et al. \cite{li2025graph} & Object graph matching & \no & \yes & \no & \no & \no & \no \\
SWA-PF \cite{chen2025swapf} & Semantic-weighted particle filter & \yes & \no & PF & class weights & partial & \no \\
\textbf{\sas} & Raster, vector, and graph semantics & \yes & \yes & proposed & explicit & calibrated & explicit \\
\bottomrule
\end{tabularx}
\end{table*}

Table~\ref{tab:comparison} clarifies that the claimed novelty is not any individual component. It is the integrated probabilistic role assigned to each component: raster evidence proposes and refines poses; object relations verify structure; persistence weights determine evidential reliability; and an integrity model decides whether the result may enter the navigation estimator.

\section{Problem Formulation}
Let the map contain semantic raster layers $M^c(\mathbf r)$, vector entities, and a graph $G_M=(V_M,E_M)$ for classes $c\in\mathcal C$. A UAV observation window produces a local bird's-eye semantic map $L_t^c(\mathbf r)$ and graph $G_t$. The candidate map-relative pose is
\begin{equation}
\boldsymbol\xi_t=[x_t,y_t,\psi_t,s_t]^T,
\end{equation}
where $s_t$ optionally absorbs residual scale uncertainty. The objective is not merely to rank tiles but to estimate a posterior
\begin{equation}
p(\boldsymbol\xi_t\mid L_{1:t},G_{1:t},U_{1:t},\mathcal M)
\end{equation}
and determine whether its best mode is safe to use as an absolute correction.

For class $c$ and local cell $\mathbf r$, the observation state is
\begin{equation}
z^c(\mathbf r)\in\{\mathrm{present},\mathrm{absent},\mathrm{unknown}\}.
\end{equation}
Unknown means that occlusion, truncation, or perception uncertainty prevents a conclusion; it must not be counted as negative evidence.

\section{SASGeo Method}
\subsection{Semantic BEV and temporal accumulation}
Camera semantic predictions are projected to a local ground plane using calibration, attitude, altitude, and optionally a DEM; road-geometry calibration can estimate such BEV homographies from lanes, borders, and crosswalks in oblique UAV video \cite{popov2026trafficcalibration}. Multiple frames are warped by VIO and accumulated:
\begin{equation}
L_t^c(\mathbf r)=\frac{\sum_{\tau=t-K}^t \omega_\tau(\mathbf r)\,\mathcal W(S_\tau^c,T_{t\leftarrow\tau})}{\sum_{\tau=t-K}^t\omega_\tau(\mathbf r)+\epsilon},
\end{equation}
where $\omega_\tau$ reflects perception and projection uncertainty.

\subsection{Stability and distinctiveness}
Persistence and geographic informativeness are separated. For feature $i$,
\begin{equation}
\sigma_i=p_{c_i}\exp(-\lambda_{c_i}a_i)d_{\mathrm{season}}(c_i,s_i)v_iq_i,
\end{equation}
where $p_c$ is a class prior, $a_i$ is map age, $\lambda_c$ is a class-specific decay rate, $d_{\mathrm{season}}$ is seasonal compatibility, $v_i$ is visibility, and $q_i$ is perception confidence. Distinctiveness is
\begin{equation}
\nu_i=-\log(P(\text{similar configuration to }i)+\epsilon),
\end{equation}
and the final evidence weight is $w_i=\sigma_i\nu_i$. A straight road can therefore be highly persistent but weakly distinctive; a bridge near a rare junction can be both.

\subsection{Three-state semantic raster likelihood}
A compatibility function distinguishes agreement, contradiction, and missing evidence:
\begin{equation}
\ell(z,m)=\begin{cases}
+\lambda_c^+, & z=\mathrm{present},\ m=\mathrm{present},\\
-\lambda_c^-, & z=\mathrm{present},\ m=\mathrm{absent},\\
-\bar\lambda_c^-, & z=\mathrm{absent},\ m=\mathrm{present},\\
0, & z=\mathrm{unknown}.
\end{cases}
\end{equation}
The raster objective is
\begin{equation}
J_R(\boldsymbol\xi)=\sum_{c,\mathbf r}w^c(\mathbf r)\ell\!\left(z^c(\mathbf r),M^c(T_{\boldsymbol\xi}\mathbf r)\right).
\end{equation}
Distance-transform compatibility is used for thin lines so small map-width errors do not dominate.

\subsection{Relational graph verification}
Nodes encode intersections, building clusters, river bends, crossings, and polygons. Edges encode relative distance, bearing, adjacency, connectivity, and crossing. Given correspondence hypothesis $\pi$,
\begin{equation}
J_G(\boldsymbol\xi,\pi)=\sum_{i\leftrightarrow j}w_i\phi_V(i,j)+\sum_{(i,k)\leftrightarrow(j,l)}w_{ik}\phi_E((i,k),(j,l)).
\end{equation}
Raster matching efficiently generates pose candidates; graph matching suppresses candidates with similar occupancy but incompatible topology.

\subsection{Temporal consistency and joint score}
A VIO increment $\Delta\hat T_t$ induces
\begin{equation}
J_T(\boldsymbol\xi_t)=-\|\operatorname{Log}(\boldsymbol\xi_{t-1}^{-1}\boldsymbol\xi_t\Delta\hat T_t^{-1})\|_{\Sigma_T^{-1}}^2.
\end{equation}
The joint objective is
\begin{equation}
J=\alpha_tJ_R+\beta_tJ_G+\gamma_tJ_T-\eta J_{\mathrm{contradiction}},
\end{equation}
with evidence-adaptive weights rather than fixed constants.

\begin{table*}[t]
\caption{Scope of the current synthetic implementation relative to the proposed formulation. This table is included to make clear which elements are experimentally exercised and which remain engineering targets for the real-world study.}
\label{tab:scope}
\centering\footnotesize
\setlength{\tabcolsep}{4pt}
\begin{tabularx}{\textwidth}{p{2.4cm}X X}
\toprule
Component & Proposed formulation & Synthetic proof of concept \\
\midrule
Semantic BEV, Eq. (4) & Multi-frame projection from camera semantics using calibration, attitude, altitude, and VIO. & Starts from synthetic semantic layers; no raw imagery, segmentation model, projection error, or VIO drift. \\
Stability, Eq. (5) & Class prior, map age, seasonal compatibility, visibility, and perception confidence. & Uses hand-set class weights; no longitudinal calibration of age or season. \\
Distinctiveness, Eq. (6) & Geographic rarity of local semantic configurations. & Hard decoys are chosen by nearest global semantic descriptors; no learned rarity model. \\
Raster likelihood, Eqs. (7)--(8) & Three-state positive, contradictory, and unknown evidence with semantic weights. & Distance-transform raster score with contradiction penalty; unknown mask removes unobserved cells from contradiction. \\
Graph term, Eq. (9) & Correspondence-hypothesis graph matching over objects and relations. & Connected-component counts and rotation-invariant pairwise-distance histograms. \\
Temporal term, Eq. (10) & Multi-frame VIO consistency and filter/factor-graph fusion. & Not evaluated; each retrieval trial is independent. \\
Integrity, Eqs. (12)--(14) & Calibrated false-fix probability and cost-sensitive acceptance. & Candidate-score margin is used as an uncalibrated rejection statistic for risk--coverage curves. \\
\bottomrule
\end{tabularx}
\end{table*}

\section{Integrity-Aware Pose Acceptance}
A false fix is a candidate whose error exceeds mission tolerance $e_{\max}$. Let $F_t$ denote this event. Accepting and rejecting incur
\begin{align}
\mathcal C_{\mathrm{accept}}&=C_{\mathrm{FF}}P(F_t=1\mid E_t),\\
\mathcal C_{\mathrm{reject}}&=C_{\mathrm{MR}}P(F_t=0\mid E_t).
\end{align}
The fix is accepted only when $\mathcal C_{\mathrm{accept}}<\mathcal C_{\mathrm{reject}}$, equivalently
\begin{equation}
P(F_t=1\mid E_t)<\frac{C_{\mathrm{MR}}}{C_{\mathrm{FF}}+C_{\mathrm{MR}}}.
\end{equation}
The probability can be calibrated from candidate margin, posterior entropy, cross-class agreement, graph consistency, unknown-evidence fraction, and VIO residual. The operating point is reported using a risk--coverage curve, where coverage is the fraction of accepted fixes and risk is the false-fix rate among accepted fixes.

\section{Synthetic Proof of Concept}
\subsection{Purpose and protocol}
The experiment asks whether structured semantic geometry can retrieve a transformed local observation from hard semantic decoys. It does not model real aerial image segmentation or closed-loop flight.

We procedurally generated 30 semantic regions of $256\times256$ cells with five layers: major roads, buildings, water, minor roads, and field boundaries. Candidate crops measured $96\times96$ cells. Each query was transformed by a randomly selected rotation in $\{0,30,60,90\}^{\circ}$, scale in $\{0.85,1.0,1.15\}$, retained crop fraction in $\{0.70,0.85,1.0\}$, rectangular occlusion up to 24\%, and class-dependent map-change probability up to 0.32. Decoys were selected from the nearest semantic descriptors rather than uniformly, making them intentionally confusable. We ran 220 trials with ten candidates each using seed 42; the released script accepts additional seeds for longer aggregate runs.

Compared variants were: (1) global semantic descriptor; (2) uniform raster distance-transform matching; (3) raster plus graph descriptor; (4) stability-weighted raster plus graph; and (5) the full model that additionally ignores unknown regions. The graph descriptor used connected-component counts and rotation-invariant pairwise-distance histograms. These simplified components test the formulation, not a production implementation.

\begin{figure}[t]
\centering\includegraphics[width=\columnwidth]{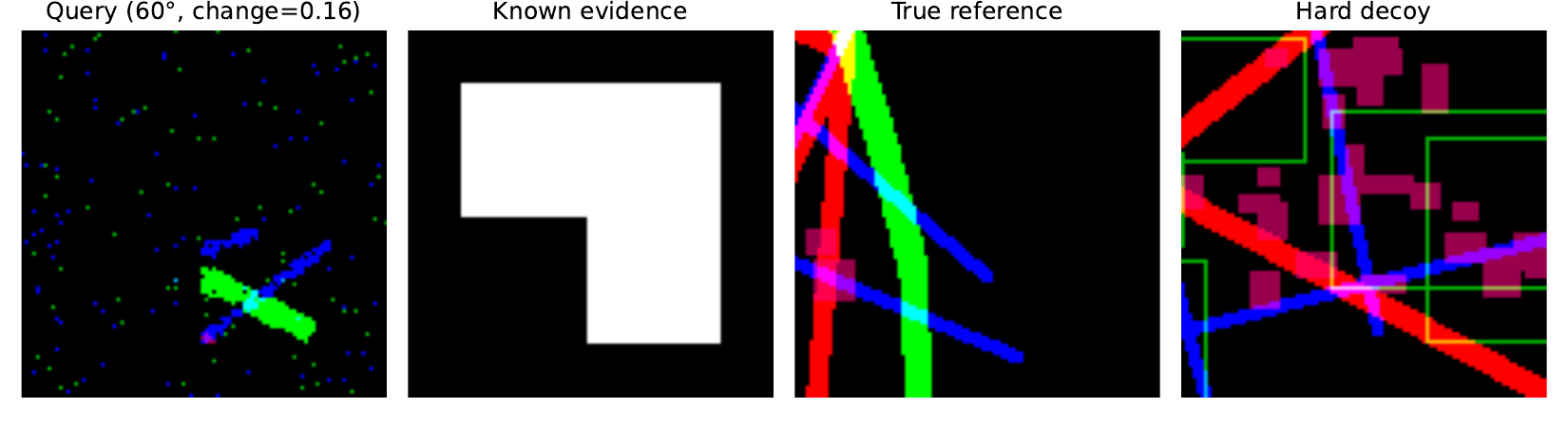}
\caption{Example synthetic query, observability mask, true reference, and hard semantic decoy.}
\label{fig:example}
\end{figure}

\subsection{Results}
\begin{table}[t]
\caption{Synthetic retrieval results over 220 randomized trials. Recall@1 intervals are Wilson 95\% confidence intervals.}
\label{tab:results}
\centering\scriptsize
\begin{tabular}{lccc}
\toprule
Method & Recall@1 (95\% CI) & Recall@5 & MRR\\
\midrule
Global descriptor & 0.586 [0.520, 0.649] & 0.886 & 0.701\\
Raster, uniform & 0.945 [0.907, 0.969] & 1.000 & 0.966\\
Raster + graph & 0.950 [0.913, 0.972] & 0.995 & 0.968\\
+ stability & 0.950 [0.913, 0.972] & 0.995 & 0.967\\
Full + unknown handling & 0.955 [0.918, 0.975] & 1.000 & 0.971\\
\bottomrule
\end{tabular}
\end{table}

\begin{figure}[t]
\centering\includegraphics[width=\columnwidth]{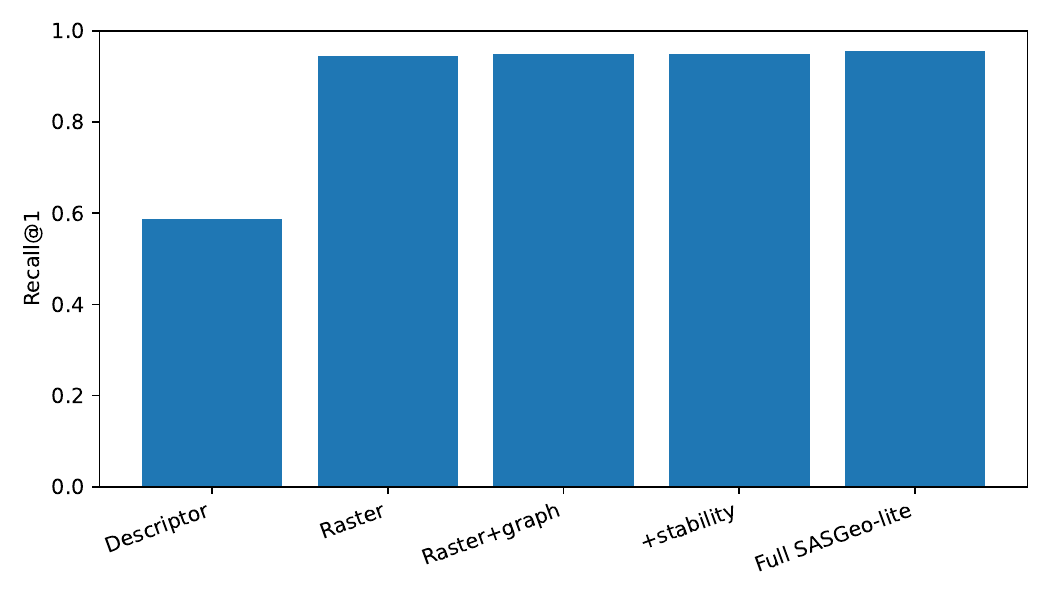}
\caption{Recall@1 ablation. Dense semantic geometry provides the dominant gain in this controlled setup; stability and unknown handling produce smaller changes.}
\label{fig:ablation}
\end{figure}

Table~\ref{tab:results} shows that a global semantic description is insufficient for hard decoys, while spatial raster alignment raises Recall@1 from 58.6\% to 94.5\%. The Wilson intervals for the descriptor and raster variants are well separated, but the intervals for the four spatial variants overlap. Therefore the supported conclusion is that dense semantic geometry is highly informative in this controlled setting; the present benchmark does not statistically distinguish graph verification, stability weighting, or unknown-evidence handling. This is a useful negative result because it exposes a ceiling effect: raster alignment is already strong on these synthetic maps, leaving too little room for the distinctive SASGeo modules to show aggregate Recall@1 gains. Targeted aliasing tests with repetitive topology, realistic map aging, and absent-match cases are required before claiming that the stability or graph terms improve localization reliability.

\begin{figure}[t]
\centering\includegraphics[width=\columnwidth]{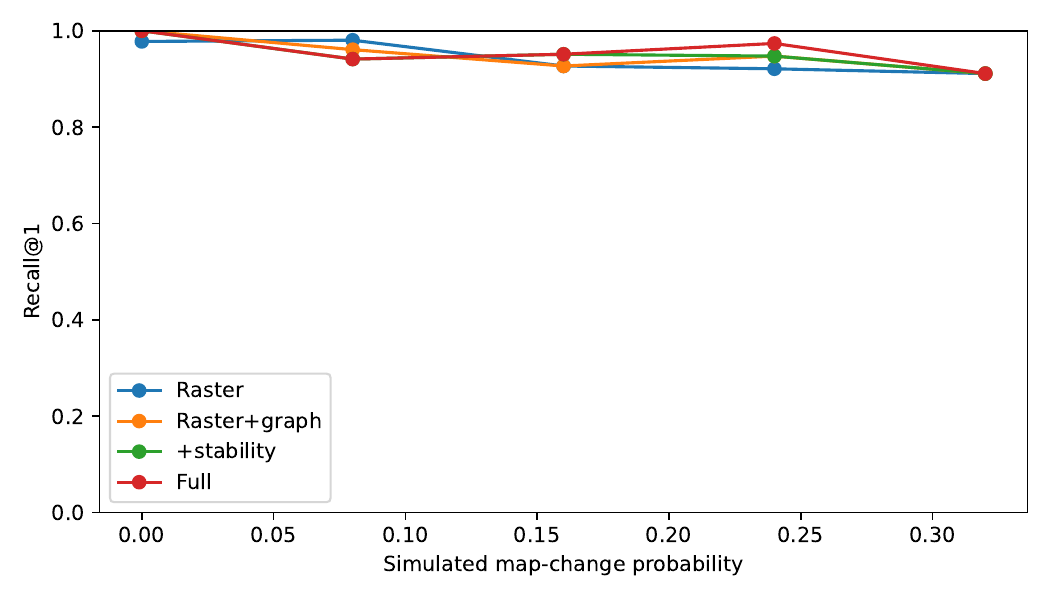}
\caption{Retrieval robustness versus simulated map-change probability.}
\label{fig:robustness}
\end{figure}

\begin{figure}[t]
\centering\includegraphics[width=\columnwidth]{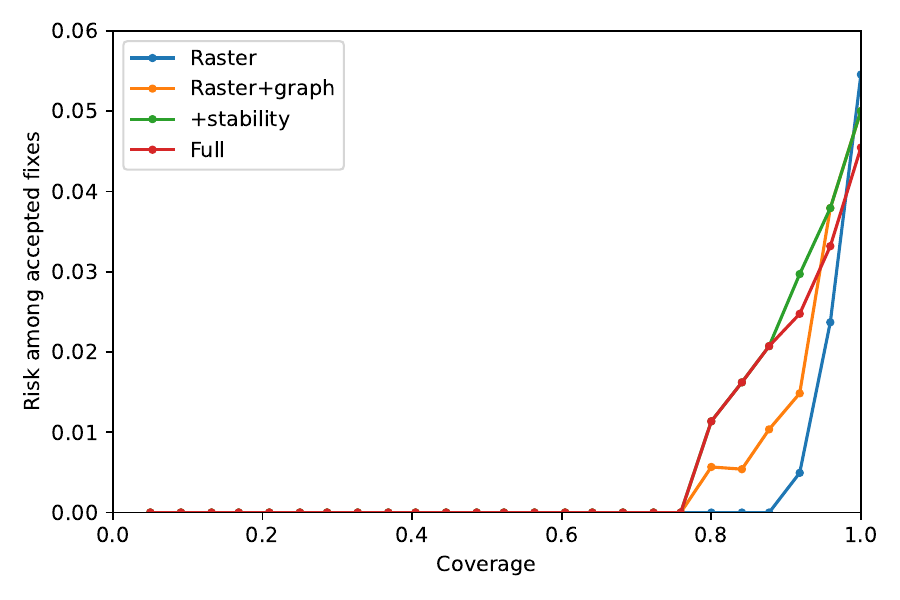}
\caption{Risk--coverage curves using candidate-score margin as an uncalibrated rejection statistic. Curves are shown for all spatial ablations to test whether the added terms reduce confident false fixes.}
\label{fig:risk}
\end{figure}

The experiment therefore validates the feasibility of the scoring pipeline and motivates a second-stage study using OSM vectors and public aerial imagery, where map age, seasonal mismatch, segmentation confidence, and true unknown regions can be measured rather than simulated.

\section{Embedded Implementation Targets}
The intended architecture separates a certified flight controller from a companion computer. A Jetson Orin NX-class module is a plausible target, but the following are design budgets, not measurements: VIO at camera rate; semantic segmentation and BEV accumulation at 5--10 Hz; coarse retrieval at 1--2 Hz; raster alignment over a small top-$N$ set; graph verification on the best candidates; and absolute correction only after multi-frame confirmation. The map pyramid stores low-resolution semantic rasters for search and compact vectors/graphs for refinement. Hybrid telemetry can preserve low-rate context while transmitting selected high-detail semantic ROIs under bandwidth limits \cite{trukhina2026hybridtelemetry}; \sas{} treats this as an implementation layer. Real profiling must report latency, memory, energy, map size, communication load, and correction frequency.

\section{Real-World Evaluation Plan}
A reproducible next step uses OSM vectors as reference semantics and publicly distributable aerial imagery as simulated UAV observations across urban, suburban, industrial, river-side, and agricultural scenes. Evaluation must use geographic holdout regions and transformations in scale, rotation, crop, occlusion, season, and map age. Besides broad retrieval, the benchmark should include targeted aliasing: repetitive road grids, similar parcels, structurally similar intersections, absent-match queries, and map-age conflicts where raster occupancy alone is ambiguous. Required metrics include Recall@1/5, metric translation and yaw error, false-fix probability, risk--coverage area, latency, memory, and map storage. Baselines should include appearance retrieval, global semantic descriptors, raster-only matching, graph-only matching, and standard sequential filtering. Real UAV sequences are ultimately required for oblique views, motion blur, VIO coupling, and navigation recovery.

\section{Limitations}
The current experiment begins from semantic layers rather than raw images and therefore does not measure perception errors. It uses discrete rotations, planar maps, simplified graphs, single-frame retrieval trials, and synthetic changes. The stability parameters are hand-set rather than calibrated from longitudinal geographic data. The confidence intervals quantify binomial uncertainty for the released 220-trial seed, not multi-region, multi-seed performance. The gains do not establish real-time feasibility or safe closed-loop operation. Feature-poor environments such as forests, deserts, snow fields, and open water may remain weakly observable.

\section{Conclusion}
\sas{} reframes UAV geo-localization as integrity-aware matching of persistent geographic structure rather than appearance retrieval alone. The proposed method integrates semantic rasters, relational graphs, temporal motion, persistence, unknown evidence, and explicit rejection. A controlled proof of concept shows that spatial semantic alignment strongly outperforms global semantic description under hard decoys. It also reveals the current ceiling effect: the benchmark validates semantic geometry but does not yet validate the distinctive persistence and graph claims. The resulting preprint is a concrete, falsifiable starting point for OSM-to-aerial and real-flight evaluation rather than a claim of a completed navigation system.

\section*{Reproducibility Statement}
The source package includes the generator, transformations, scoring functions, fixed seed, raw trials, summary metrics, figure code, and OSM vector extract for Fig.~\ref{fig:intuition}. The complete source code is publicly available at \url{https://gitlab.com/emilab-group/sasgeo}. It avoids nonstandard compiled image-processing dependencies and regenerates artifacts under \texttt{figures/} and \texttt{results/}. Run \texttt{python run\_synthetic\_experiment.py}; longer checks can use \texttt{--seeds 42,43,44,45,46}.

\end{document}